\documentclass{article}

\usepackage[preprint]{neurips_2021}

\usepackage[utf8]{inputenc} 
\usepackage[T1]{fontenc}    
\usepackage{hyperref}       
\usepackage{url}            
\usepackage{booktabs}       
\usepackage{amsfonts}       
\usepackage{nicefrac}       
\usepackage{microtype}      
\usepackage{xcolor}         
\usepackage{graphicx}  

\usepackage{caption}
\usepackage{subcaption}

\title{ClipMatrix: Text-controlled Creation of 3D Textured Meshes}

\author{%
  Nikolay Jetchev, \\
  Zalando Research, Zalando SE, Berlin \\
  \texttt{nikolay.jetchev@zalando.de} \\
}

\begin{document}

\maketitle
\vspace{-0.5cm}

\begin{abstract}
If a picture is worth thousand words, a moving 3d shape must be worth a million.
We build upon the success of recent generative methods that create images fitting the semantics of a text prompt, and extend it to the controlled generation of 3d objects. 
  We present a novel algorithm for the creation of textured 3d meshes, controlled by text prompts. Our method creates aesthetically pleasing high resolution articulated 3d meshes, and opens new possibilities for automation and AI control of 3d assets.   
  We call it "ClipMatrix" because it leverages CLIP text embeddings to breed new digital 3d creatures, a nod to the Latin meaning of the word "matrix" - "mother".
  See the online \href{https://twitter.com/NJetchev/}{\color{blue}gallery} for a full impression of our method's capability.
\end{abstract}

\begin{figure} [h] \centering 
\includegraphics[width=4.5cm]{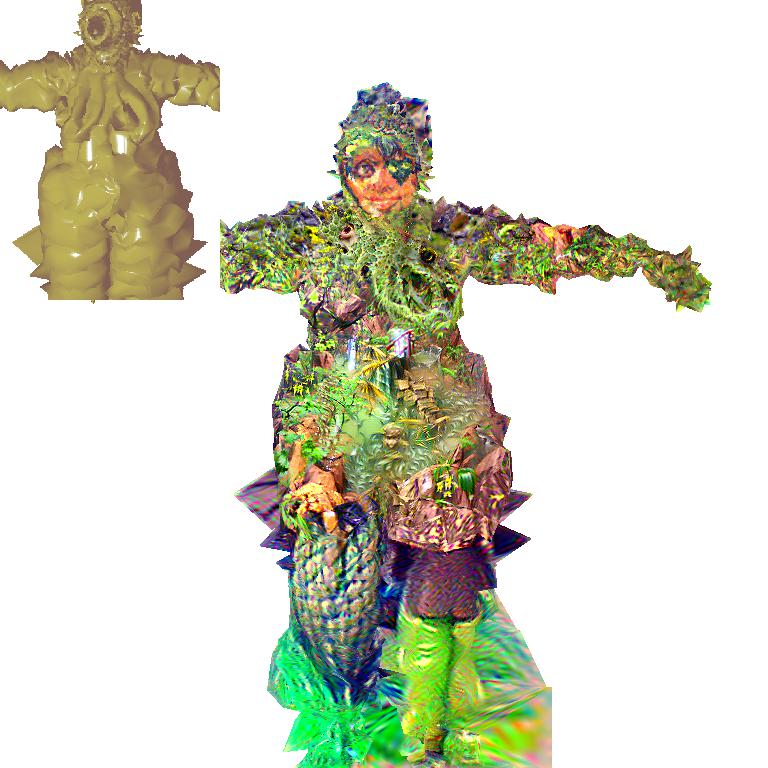}
\includegraphics[width=4.5cm]{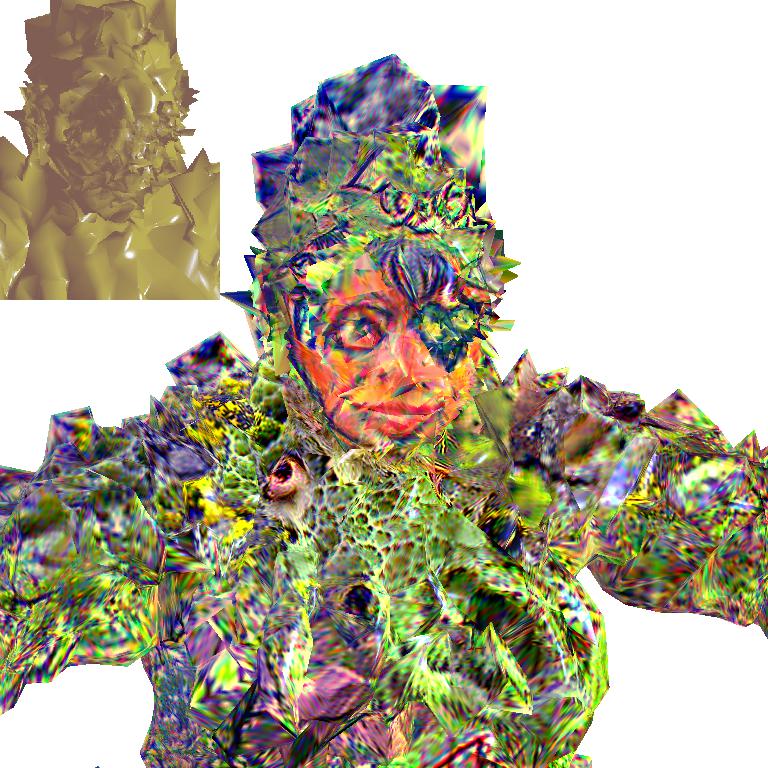}
\caption{Example of ClipMatrix capabilities: a 3d textured model is optimized w.r.t. the prompt "Green witch of the swampy organic ocean with tentacles", and the 3d mesh (without texture) w.r.t. the prompt "suction valves and tentacles". Two snapshots show the final result rendered from different cameras. The real beauty of 3d art is revealed in motion of limbs and camera - see \href{https://twitter.com/NJetchev/status/1438129645233315843}{\color{blue}video}. For computation we did 600 loss gradient optimization steps, taking 20 minutes on a Nvidia P100 GPU.
} 
\end{figure}

\vspace{-0.5cm}

\section{ClipMatrix: Background and Method}
Pretrained neural networks know a lot about the visual world - and visualizing their learned representations as images is a digital artform with a passionate online following. Approaches creating 2d images as output are everywhere on the net, due to their instant appeal -- colourful aesthetics, fast to train, easy to  modify, suitable for social network sharing.
Deepdream and related differentiable image parametrisations \cite{mordvintsev2018} were among the first approaches to show how optimizing neural networks w.r.t. input pixels can lead to beautiful art.
More recently, CLIP \cite{radford2021learning} ushered a new era for generative art -- its joint embedding space $\phi$ relates both image and text modalities, which allows artists and ML practitioners to flexibly play with both. Telling the AI "draw me object X" and then the AI draws "X" is a powerful creativity paradigm. Many artists and researchers \cite{aleph} showed what beauty can arise by optimizing image similarity with a text embedding.
The CLIP representations are so flexible that they can guide also the creation of 3d graphics.
CLIP is already used for 3d learning \cite{jain2021putting} with an image reconstruction objective. However, this rigid supervision limits artistic creativity; also NeRF fitting has huge computational cost.

In contrast, ClipMatrix is build around performant high-resolution mesh models as 3d representation. Our method can surprise the user with novel shapes and textures. 
ClipMatrix is controlled by the semantic similarity to CLIP's text embeddings - different objective with many more optima than reconstruction supervised loss.
As initial mesh we use a parametric rigged human body model \cite{smplx}. 
ClipMatrix tunes these parameters: 

\begin{itemize}
    \item $\beta$ SMPL body shape
    \item $\theta$ joint pose of the rigged SMPL model
    \item $\delta$ deformation per SMPL vertex
    \item $x$ texture image
    \item $c,l,m$ camera, light and material parameters
\end{itemize}

\begin{figure} [t] \centering 
\includegraphics[height=2.3cm]{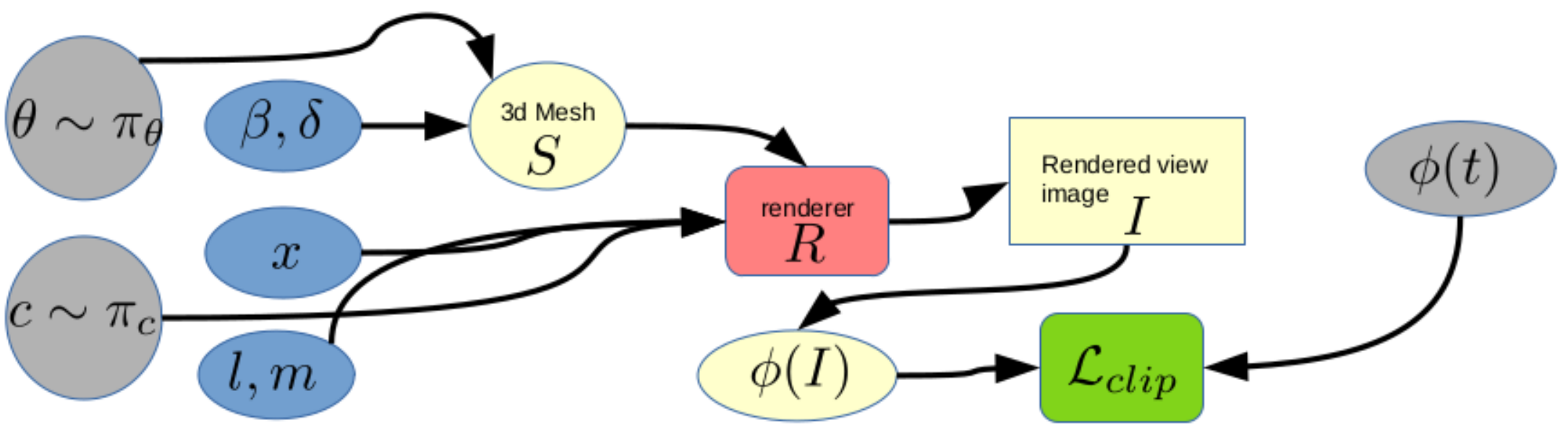}
\caption{Schema of ClipMatrix:  parameters $\beta,\delta,x,l,m$ (SMPL shape, vertex deform, texture, light and material) are optimized; random camera views and body poses $c,\theta$ are sampled.
All these influence the renderer $R$, which creates the final  2d image views $I$. These views are embedded in CLIP space $\phi(I)$, and used together with input text prompts $\phi(t)$ in a loss $\mathcal{L}_{clip}$. We show this for one prompt $t$ only, but in general multiple prompts ca be used to define loss sum terms.} 
\label{fig:schema}
\end{figure}

The final rendered image output is $I= R(S(\beta,\theta,\delta),l,m,c,x)$, see Fig. \ref{fig:schema}.
Here $S$ is the mesh output from SMPL (given the mesh params); $I$ is the rendered image given mesh, camera, material, light and texture.
We leverage Pytorch3d \cite{ravi2020pytorch3d} as a performant differentiable 3d renderer $R$. 
ClipMatrix connects images of rendered 3d views and text prompts in a fully differentiable loss function. We sample camera $c$ and pose $\theta$, and minimize the expected loss w.r.t. the parameters:

\vspace{-0.3cm}
\begin{eqnarray}
\mathcal{L}(\beta,\delta,x,l,m) = \sum_{t}  \mathop{\mathbb{E}}_{\theta \sim \pi_{\theta},c \sim \pi_c} \mathcal{L}_{clip}(I,t) + \lambda \mathcal{L}_{reg}(S)
\end{eqnarray}
\vspace{-0.3cm}

\begin{figure} [h] \centering 
\includegraphics[width=13cm]{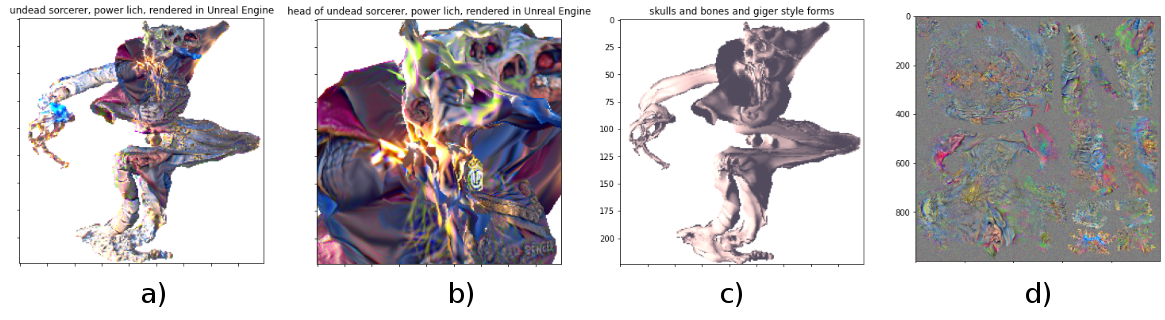}
\caption{Illustration how ClipMatrix couples 3d mesh views with text control. Text prompts are shown on top of images. (a,b,c) different rendering images used for a set $\{t_0,t_1,t_2 \}$ of text prompts, enabling  enhanced control of the final results. (d) the learned UV texture is used for the renders (a,b) but not for (c) which uses plain material. W.l.o.g. we can have unique textures and cameras for each prompt - e.g. (b) zooms-in on the creature head, and the prompt says "head of undead sorcerer."} 
\label{fig:texts}
\end{figure}

By sampling random camera $c \sim \pi_c$ we ensure our output mesh has the desired properties from any viewing angle. In contrast, optimizing a single fixed camera $c^*$ makes a method for simpler 2d image generation. 
Similarly, we sample random poses $\theta \sim \pi_{\theta}$ to leverage the dynamism of the rigged 3d model, as opposed to a static sculpture.
$\mathcal{L}_{reg}$ is a standard 3d mesh regularization \cite{mir20pix2surf} weighted by $\lambda$, keeping deformed meshes 'well-behaved'.
$\mathcal{L}_{clip}(I,t) = -cos(\phi(I),\phi(t))$ is the negative cosine similarity in CLIP embedding space $\phi$ between image $I$ and the embedding of the fixed input text prompt $t$, as used by \cite{aleph}. We can flexibly sum over multiple text prompts $\{ t_i \}$. 
In addition, we use specifically defined camera distributions $\pi_c$ to enabling  specific meshpart-text correspondence, e.g. Fig. \ref{fig:texts})b) samples a grid of cameras centered around the mesh head.

\section{Summary and Outlook}
We presented ClipMatrix: a novel generative art tool that allows the text-controllable creation of high resolution 3d textured shapes. The method leverages the SMPL mesh model with a CLIP loss. 
The framework is very flexible, and practitioners can get a range of appealing results when engineering different text prompts and camera views. Appendix I and the online gallery  \href{https://twitter.com/NJetchev/}{\color{blue}gallery} showcase sample creations.
As a limitation, we note that optimization of discrete mesh parameters is quite sensitive to tweaks of the learning rate and regularisation strength $\lambda$. While acceptable for curated generation,  this instability currently prevents fully automated 3d asset creation.
We plan to investigate other 3d parametrisations like implicit surfaces - they can improve stability, but are costly in terms of image resolution and computational speed.

\clearpage
\newpage

\bibliographystyle{unsrtnat}
\bibliography{bibi}

\section*{Ethical Implications}
We see no specific risks related to the  current work that exceeds the risks of similar 2d generation approaches. ClipMatrix is a tool allowing playful exploration and novel creation for artists. Such art does not touch any critical issues, such as privacy and personal data. It is also a tool requiring a human-machine interaction for best results (exploration, curation, quality control), so no full automation is possible yet.
Full automation of 3d asset cr[preprint]eation will ultimately be disruptive to the 3d modelling and animation industry, but we don't see this happen in the foreseeable future.

\section*{Appendix I: Additional Results}
\label{app1}
\begin{figure}
     \centering
     \begin{subfigure}[h]{0.41\textwidth}
         \centering
         \includegraphics[width=\textwidth]{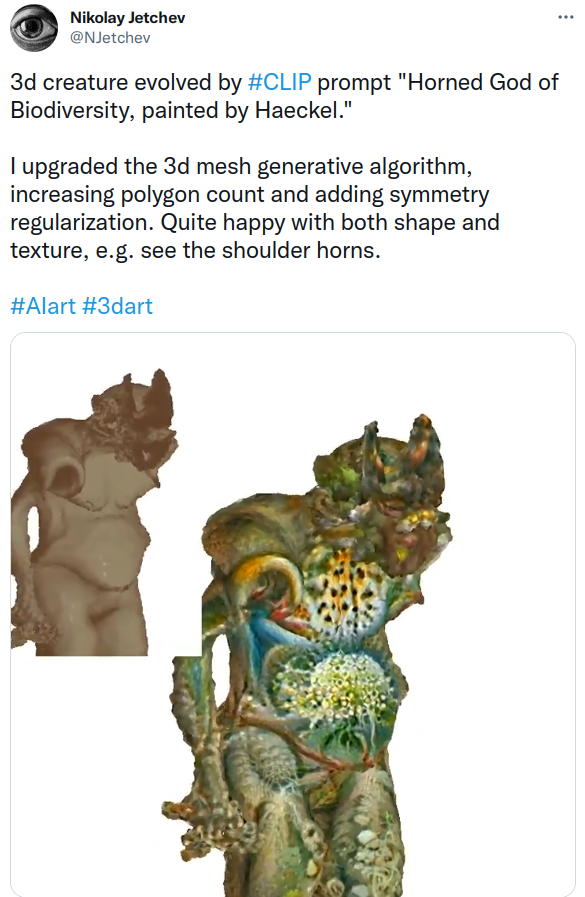}
         \caption{\href{https://twitter.com/NJetchev/status/1440432169768218625}{\color{blue}video}}
     \end{subfigure}
     \hfill
     \begin{subfigure}[h]{0.41\textwidth}
         \centering
         \includegraphics[width=\textwidth]{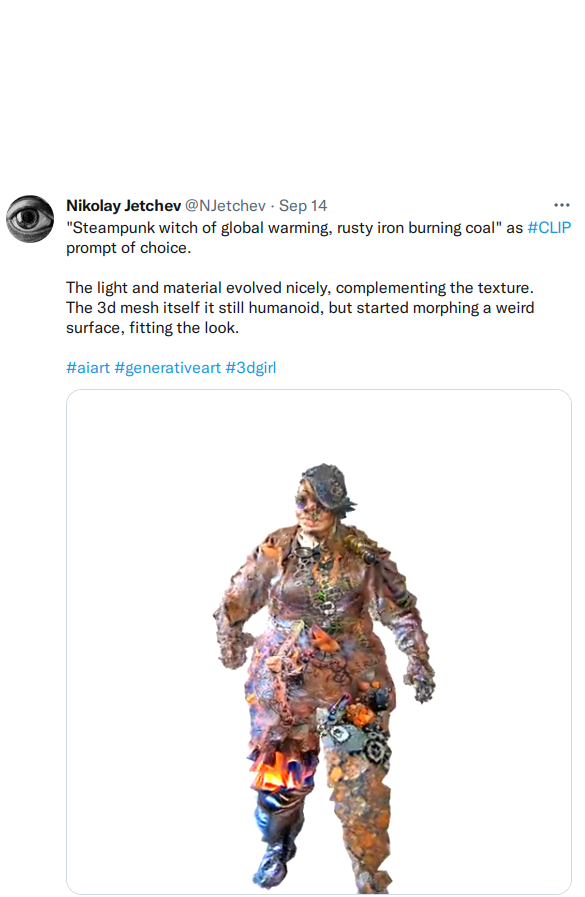}
         \caption{\href{https://twitter.com/NJetchev/status/1437804396302278665
}{\color{blue}video}}
     \end{subfigure}
     \hfill
     \begin{subfigure}[h]{0.41\textwidth}
         \centering
         \includegraphics[width=\textwidth]{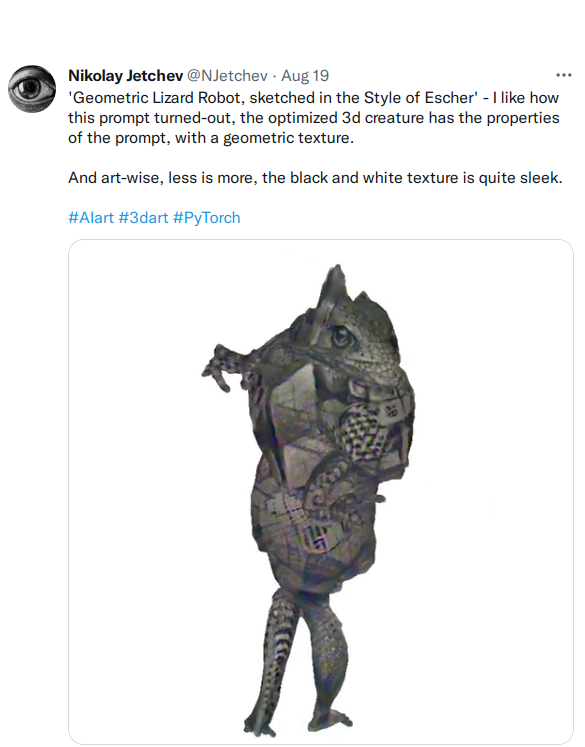}
         \caption{\href{https://twitter.com/NJetchev/status/1428435305120649220}{\color{blue}video}}
     \end{subfigure}
     \hfill
     \begin{subfigure}[h]{0.41\textwidth}
         \centering
         \includegraphics[width=\textwidth]{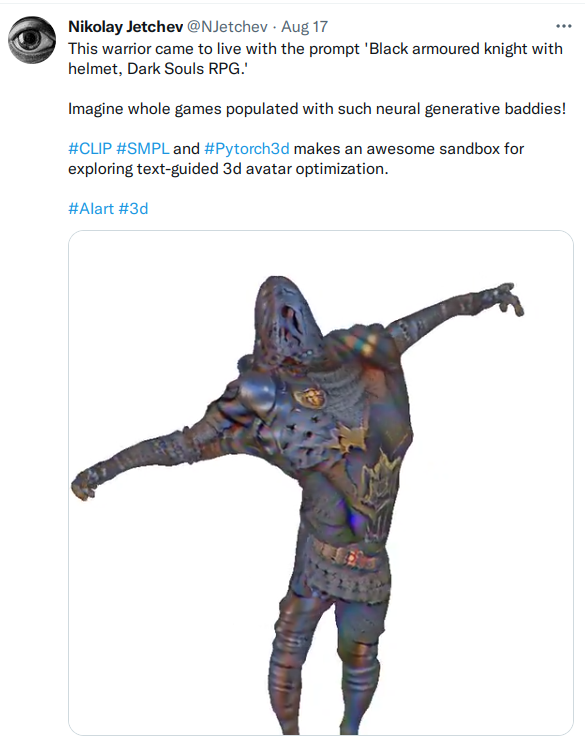}
         \caption{\href{https://twitter.com/NJetchev/status/1427740192631402499}{\color{blue}video}}
     \end{subfigure}
        \caption{Examples of ClipMatrix 3d artwork. 
        See tweet text above each image for a description of the unique design choices explored inside each artwork.
        Click the video links for an animated viewing experience: a rotating camera and body pose interpolation shows different facets of each artwork.}
        \label{fig:snap}
\end{figure}

While creating art via optimization sounds straightforward, the design of loss function and rendering priors is a long process of trial and error experimentation. The ClipMatrix framework can produce many different results depending on the design choices. This includes the degree of penalizing mesh deviation from the base human form, how much to allow lighting and material to differ, how to place cameras and how many unique CLIP prompts to use as sum terms in the loss definition, etc. Figure \ref{fig:snap} shows four examples (out of many more available online) of the evolution of the ClipMatrix method. Each of these presents a step in the improvement of the method, as the tweets and timestamps of the artworks indicate.
We expect the method to change even more in the future, and would be very happy if the users contact the authors and share ideas for technical improvements, or interesting text prompts and sample artwork.

\section*{Appendix II: Technical Details}
We use the SMPLx model \cite{smplx} as underlying mesh.
It has around 10000 vertices, and 20000 triangle faces.
We render the (textured) mesh at size 224x224 pixels, which is also the image size for CLIP embeddings.
For inference and video post-processing, we typically render at size 768x768 pixels. Since this is a 3d model, output size can be flexibly adjusted depending on the context and model details.
We optimize a texture $x$ of size 1024x1024 pixels, corresponding to the SMPLx UV coordinates.
With the 224x224 render size, we fit 4 random camera views per minibatch for training, on a 16GB GPU card. Given the complexity of the overall rendering pipeline, a lot of tweaks are possible between image quality and memory computation footprint.

\end{document}